%% file: ms.tex
\documentclass[sigconf]{acmart}

\input{macros}

\AtBeginDocument{%
  \providecommand\BibTeX{{%
    \normalfont B\kern-0.5em{\scshape i\kern-0.25em b}\kern-0.8em\TeX}}}

\copyrightyear{2019}
\acmYear{2019}
\setcopyright{acmlicensed}
\acmConference[ACSAC 2019]{2019 Annual Computer Security Applications Conference}{December 9--13, 2019}{San Juan, Puerto Rico, USA}
\acmBooktitle{2019 Annual Computer Security Applications, December 9--13, 2019, San Juan, Puerto Rico, USA}

\begin{document}

%
\title[\toolName: Multiple-model-based Defense]{\toolName: Multi-model-based Defense Against Adversarial Examples for Neural Networks}

\author{Siwakorn Srisakaokul}
\affiliation{\institution{University of Illinois at Urbana-Champaign}}
\email{srisaka2@illinois.edu}

\author{Yuhao Zhang}
\affiliation{\institution{Peking University}}
\email{zhang_yuhao@pku.edu.cn}

\author{Zexuan Zhong}
\affiliation{\institution{University of Illinois at Urbana-Champaign}}
\email{zexuan2@illinois.edu}

\author{Wei Yang}
\affiliation{\institution{University of Texas at Dallas}}
\email{wei.yang@utdallas.edu}

\author{Tao Xie}
\affiliation{\institution{University of Illinois at Urbana-Champaign}}
\email{taoxie@illinois.edu}

\author{Bo Li}
\affiliation{\institution{University of Illinois at Urbana-Champaign}}
\email{lbo@illinois.edu}

%
\renewcommand{\shortauthors}{Trovato and Tobin, et al.}

%
\begin{abstract}
Despite being popularly used in many applications, neural network models have been found to be vulnerable to \textit{adversarial examples},  i.e., carefully crafted examples aiming to mislead machine learning models.
Adversarial examples can pose potential risks on safety and security critical applications.
However, existing defense approaches are still vulnerable to 
attacks, especially in a white-box attack scenario. To address this issue, we propose a new defense approach, named \toolName, based on \textit{robustness diversity}. Our approach consists of (1) a general defense framework based on multiple models and (2) a technique for generating these multiple models to achieve high defense capability.
In particular, given a target model, our framework includes multiple models (constructed from the target model) to form a model family. The model family is designed to achieve robustness diversity (i.e., an adversarial example successfully attacking one model cannot succeed in attacking other models in the family). At runtime, a model is randomly selected from the family to be applied on each input example. Our general framework can inspire rich future research to construct a desirable model family achieving higher robustness diversity. Our evaluation results show that \toolName (with only up to 5 models in the family) can substantially improve the target model's accuracy on adversarial examples by 22--74\% in a white-box attack scenario, while maintaining similar accuracy on legitimate examples.
\end{abstract}

%
%


%

%
\maketitle

\input{introduction}
\input{background}
\input{approach}
\input{evaluationsetup}
\input{evaluationresult}

\input{discussion}
\input{relatedwork}

\input{conclusion}

%
\bibliographystyle{ACM-Reference-Format}
\bibliography{ms}

%
\appendix
\section{Datasets}
Here is the description of the two popular public datasets that we use:\\
\noindent \textbf{MNIST} is a dataset of handwritten digits, consisting of ten labels for the ten digits. We select 60,000 examples for the training set and 10,000 examples for the test set. Each image is a 28x28 black and white image.

\noindent \textbf{CIFAR-10} is a widely used dataset consisting of 10 labels. We select 50,000 examples for the training set and 10,000 examples for the test set. So there are 6,000 images per class. Each image is a 32x32 color image.

\end{document}

%% file: macros.tex
\usepackage{xcolor}
\usepackage{xspace}
\usepackage{amssymb}
\usepackage{hyperref}
\usepackage{graphicx}
\usepackage{subcaption}
\usepackage{array}
\usepackage{color}
\usepackage{colortbl}
\usepackage{amssymb}
\usepackage{pifont}
\usepackage{multirow}

\newcommand{\toolName}{\textsc{MulDef}\xspace}

%% file: introduction.tex
\section{Introduction}
\label{sec:intro}

Neural networks recently have been used to solve many real-world tasks such as image recognition and can achieve high effectiveness on these  tasks~\cite{He2016DeepRL}. However, given a legitimate example that the model can correctly classify, previous research~\cite{Carlini2017Towards,Nicolas2017Practical,Narodytska2016SimpleBA,papernot2017cleverhans} proposed various attack approaches to perturb the example by applying imperceptible modification on the example to fool the model, i.e., causing the model to misclassify the perturbed example.
We refer to such perturbed example \textit{adversarial example} and the model being attacked \textit{target model}.
These attack approaches can be used in two attack scenarios: (1) white-box attack scenario where the attackers have complete knowledge about the target model (and also its defense approaches),
and (2) black-box attack scenario where the attackers do not know anything about the target model (or its defense approaches), but know the output produced by the model, given an arbitrary example. We focus on improving the model against the white-box attack scenario, which is known to be harder to defend against.

With various effective attack approaches being invented for the two attack scenarios, a number of defense approaches such as adversarial training~\cite{goodfellow2014explaining,Carlini2017Towards} and defensive distillation~\cite{Nicolas2016Distillation} were proposed. However, these existing approaches are not really effective, especially in a white-box attack scenario, facing three main limitations.
(1) \textit{Ineffectiveness against re-attack}.
 The improved target model resulted from some defense approaches such as adversarial training is still vulnerable to new adversarial examples generated by reapplying the same attack approach on the improved model in the white-box attack scenario.
 These attack approaches rely on computing the model's gradient.
 Even after the defense approach of adversarial training improves the model with additional adversarial examples in the training set, the attack approaches can still compute the gradient of the improved model and generate new adversarial examples. One potentially effective variation of adversarial training includes modification of the loss function used to optimize the model parameters~\cite{papernot2017cleverhans}, but that variation requires manually changing the target model's implementation.
 (2) \textit{Ineffectiveness to transferable attack}.
  Adversarial examples have the \textit{transferability} property:  adversarial examples can be used to transfer attacks across models~\cite{szegedy2013}. So attackers can train a substitute model (which is white-box) for the target model and generate adversarial examples for the substitute model to indirectly attack the target model~\cite{Carlini2017Towards}.
  This transferability property can also be used to attack the target model in a black-box attack scenario~\cite{papernot2016}.
 (3) \textit{Bypassable distillation.}
 Even after defensive distillation, attackers can still compute the gradient of the inputs to the pre-softmax layer and reduce the magnitude of the inputs to the softmax layer~\cite{carliniw2016}.

To address these three main limitations, we propose a new defense approach, named \toolName, based on robustness diversity. Our approach consists of (1) a general defense framework based on multiple models and (2) a technique for generating these multiple models to achieve high defense capability. The general framework includes two components: the model generator and  runtime model selector. The design of the general framework is based on the design principle of \textit{diversity}.
Such diversity among multiple models introduces uncertainty in the target model~\cite{Larsen2014}, making it harder to attack. We design our general framework based on our main insight 
that existing attack approaches attack a single-model machine learning system by  computing the target model's gradient  based on its loss function. 

In particular, the model generator constructs a model family to assure that an adversarial example generated for one model in the family cannot fool other models in the family. This model family can address the limitations of ineffectiveness to re-attack and ineffectiveness against transferable attack. To address the limitation of bypassable distillation, the runtime model selector uses a low-cost random strategy to select a model in a model family to be applied on a given example such that the attackers do not know beforehand which model to compute the gradient even when distillation can be bypassed. This random strategy is also a burden for the reverse engineering efforts made by attack approaches. Note that generally (deterministic) runtime analysis (such as multi-model majority voting~\cite{pei17:deepxplore,srisakaokul2018:multitest}) can be conducted on the given example and all the models in the family in order to select a model that is likely to be robust to the example. However, such runtime analysis is costly and unscalable.

In our general defense framework, the runtime model selector has to select a model (in the family of diverse models) that is robust to any given example with a high chance. Thus there are two main desirable properties of the model family. (1) \textit{Legitimate-behavior preservation}. The models in the family shall preserve the same accuracy on legitimate examples as the target model. (2) \textit{Robustness diversity}. Given an adversarial example, the majority of the models in the family are robust to the example; in this way, even when the adversarial example is carefully constructed to attack one model in the family successfully (e.g., in a white-box attack scenario), there is $(N-1)/N$ chance for randomly choosing from the family another model robust to the adversarial example, where $N$ is the family size (i.e., the number of the models in the family). We introduce a new metric named \textit{diversity measurement} to measure the robustness diversity of a model family.

In addition, while aiming to satisfy the preceding two properties, $N$, the number of models generated for the family, shall aim to be as high as possible in order to to increase the defense capability. The reason is that the chance for randomly choosing the model based on which the adversarial example is carefully constructed (and thus likely successful in attacking) is $1/N$. The larger the value $N$ is, the higher the chance of choosing a robust model for a given example is. However, increasing $N$ can make satisfying the preceding two properties more challenging.

Our general framework can inspire rich future research (on both theory and systems) from the research community to construct a family of as many models as possible given the target model such that the family satisfies the preceding two properties as much as possible. There exist some open questions for future research. For example, given a technique for constructing the family models, what is the theoretically proven or empirically demonstrated bound of $N$, the number of models that can be constructed to satisfy the two properties? Given a fixed $N$, what is the theoretically proven or empirically demonstrated bound of robustness diversity while achieving the legitimate-behavior preservation? 

To demonstrate the benefits of our general framework, our approach includes a technique for generating these multiple models to achieve high defense capability. In particular, for the first property (i.e., legitimate-behavior preservation), our technique constructs each additional model by using the same architecture and parameter configuration as the target model, and the majority of the training examples are from the original training set to train the target model. For the second property (i.e., robustness diversity), the models in the family should be diverse and complementary. To accomplish so, our technique trains later-constructed models with some adversarial examples for the earlier-constructed models.

In summary, this paper makes the following main contributions:
\begin{itemize}
    \item A general defense framework based on multiple models (with high robustness diversity) constructed from  the target model.
    
    \item A novel technique for generating these multiple models to achieve high defense capability. 
    
    \item Evaluation results on three attack approaches (Fast Gradient Sign Method~\cite{papernot2017cleverhans}, Carlini \& Wagner attack~\cite{Carlini2017Towards}, Projected Gradient Descent~\cite{madry2018towards}) for showing  substantial benefits of our general framework instantiated with the technique for improving defense capability.
\end{itemize}

The evaluation results show that our defense approach substantially improves the robustness of the target model (on MNIST~\cite{mnist} and CIFAR-10~\cite{cifar10} datasets) in the case of a white-box attack scenario, and slightly improves the target model in the case of a black-box attack scenario. We implement \toolName in a tool and conduct all the evaluations in Python 3.0. We use TensorFlow 1.8.0~\cite{tensorflow} for the machine learning computation and Keras (the Python Deep Learning library version 2.1.6)~\cite{keras} for neural networks. 
All the implementations can be found on the project website\footnote{Project website: URL anonymized due to double blind submission}.

%% file: background.tex
\section{Background}
\label{sec:background}
In this section, we illustrate the terminology and basic attacks and defense approaches in previous work.
\subsection{Legitimate vs. Adversarial Examples}
We focus on addressing adversarial example attacks on neural network models for classification tasks. 
A \textit{legitimate example} $x$ is an example that occurs naturally~\cite{Meng2017MagNet} for the classification task. 
For example, if a classification task is to classify digits, legitimate examples can be images of real digits without other elements. 
An \textit{adversarial example}~\cite{Szegedy2013prop} $x'$ is an example similar to a legitimate example with imperceptible changes (of the legitimate example) that can change the target model's prediction on the example.

\subsection{Existing Attack Approaches}
An attack approach takes a legitimate example and then tries to generate an adversarial example similar to the legitimate example. 
Thus in our evaluation setup (Section~\ref{sec:evalresults}), we control each attack configuration to generate only an adversarial example within a certain distance from the given legitimate example in order to avoid human effort to label the adversarial example.

We evaluate our proposed defense approach against major existing attack approaches as described below.

\noindent \textbf{Fast Gradient Sign Method (FGSM).} 
FGSM~\cite{goodfellow2014explaining} generates adversarial examples iteratively.
Let $l$ be the ground-truth label of $x$. 
Let $J(x, l)$ be the loss function of classifying $x$ as label $l$. 
For each pixel, FGSM updates the pixel according to the sign of the gradient of the loss function at the pixel. 
Formally, FGSM iteratively modifies $x$ as follows:
$$
x' = x + \epsilon \cdot \textit{sign}(\nabla_x J(x, l))
$$

\noindent \textbf{Carlini \& Wagner attack (C\&W).} C\&W~\cite{Carlini2017Towards} generates adversarial examples with small perturbation $\delta$ through the following optimization
$$
\delta = \mathop{\arg\min}_{\delta'} D(x, x+\delta') + c \cdot f(x+\delta')
$$
where D is a distance metric, which can be $L_0$ (\# of nonzero elements), $L_2$ (euclidean distance), or $L_{\infty}$ (largest magnitude among each element), $c$ is a constant to balance constraints, and a logit-based objective function $f(\cdot)$ is designed in such a way that $f(x') \leq 0$ if and only if the classifier misclassifies $x'$, indicating that the attack succeeds. Such logic-based objective function enables C\&W to generate adversarial examples robust against the defensive distillation~\cite{Nicolas2016Distillation}.

\noindent \textbf{Projected Gradient Descent (PGD).} PGD~\cite{madry2018towards} is similar to FGSM, but more powerful. PGD is an iterative variant of FGSM. In each iteration, PGD updates the example as follows:
$$
x'_{i+1} = \pi\{FGSM(x'_{i})\}
$$
where $\pi$ is a clip function to keep $x'_{i+1}$ within a defined perturbation range.

There are different threat levels of adversarial attack, categorized by the amount of attackers' knowledge and their goal~\cite{papernot:eurosp:2016}. In a  white-box attack scenario, the attackers have complete knowledge of the target model. The attackers' goal is \textit{misclassification}, altering the output classification to any class different from the original class. A  black-box attack scenario has the same goal, but less knowledge. The attackers have access to only the \textit{oracle}. 

We focus on the white-box attack scenario; however, our evaluation also includes the black-box attack scenario.

\subsection{Existing Defense Approaches}

We next describe two main existing defense approaches proposed in previous work. 

\noindent \textbf{Adversarial training.}
The idea of adversarial training~\cite{goodfellow2014explaining,Carlini2017Towards} is to make the target model more general and have some exposure to adversarial examples.
A straightforward way is to augment the training set by replacing some training samples with the corresponding adversarial examples generated by an attack approach~\cite{moosavi2016deepfool}. 
Also, one can train the model using an adversarial objective function to improve the robustness and generality of a classifier~\cite{madry2018towards}. Adversarial examples can also be crafted from pre-trained models (e.g., ones from Ensemble Adversarial Training)~\cite{tramer2018ensemble}.

\noindent \textbf{Defensive distillation.}
Some attack approaches rely on optimizing an objective function by computing the gradient of the target model.
Thus it would be useful for a defender to hide the gradient of the target model.
Defensive distillation~\cite{Nicolas2016Distillation} trains a classifier to cause a rapid reduction of its gradient over an input,  resulting in that an attacker can hardly perform an attack requiring computing gradients.
Defensive distillation hides the gradient between the pre-softmax layer and softmax outputs by using distillation training.

Some other existing defense approaches address the problem in different ways. First, they detect whether an example is adversarial by training a model detector~\cite{Lu2017safetynet}. Second, they reform the detected adversarial example to a legitimate example~\cite{Meng2017MagNet} or classify the adversarial example by a highly nonlinear model~\cite{Krotov2017dense}.

In general, existing defense approaches extend the target model in various ways. Some approaches improve the target model against adversarial examples by modifying the weights and parameters of the target model. Some approaches~\cite{Meng2017MagNet} do not modify the target model at all, but detect and modify adversarial examples before passing them to the target model. Our proposed \toolName approach does not modify the target model or adversarial examples. We evaluate our \toolName approach on the following two metrics.
\begin{itemize}
\item \textbf{Test accuracy (TestAcc)}: the accuracy on a legitimate example in the test set.
\item \textbf{Adversarial accuracy (AdvAcc)}: the accuracy on an adversarial example generated by an attack approach. The label of an adversarial example is the label of the legitimate example used to generate that adversarial example.
\end{itemize}

%% file: approach.tex
\section{Our \toolName Approach}
\label{sec:approach}
We design our \toolName approach for improving a neural network model to invalidate the reverse engineering efforts  made by attack approaches.
The design of the \toolName approach is based on the design principle of diversity. Diversity achieved by  randomly selecting a model from a family of models introduces uncertainty in the target under defense. 

\begin{figure}[t]
    \centering
    \includegraphics[scale=0.35]{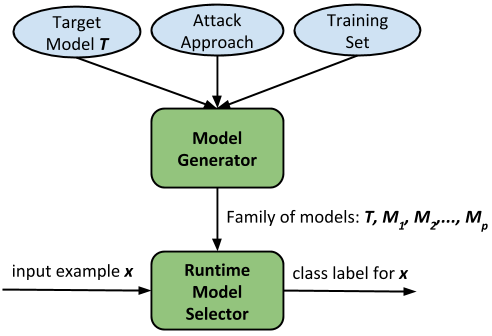}
    \caption{Our general defense framework. Model generator constructs $p$ similar models to $T$: $M_1, M_2, \dots, M_p$. Given an input example $x$, \toolName randomly selects one model to compute the class label for $x$.}

    \label{fig:toolframework}
\end{figure}

The \toolName approach includes a general defense framework as shown in Figure~\ref{fig:toolframework}. 
The framework consists of two components: (1) the model generator and (2) the runtime model selector. 
The two components work together to improve the target model's effectiveness on adversarial examples. 
As described in Section~\ref{sec:intro}, the model generator aims to produce a model family for achieving the two properties: (1) legitimate-behavior preservation and (2) robustness diversity. In addition, we aim to have as many models as possible to increase the chance that the runtime model selector ends up selecting a robust model for any given example.

First, let us formulate the problem. Given a target model $T$, we construct a model family (consisting of $T, M_1, M_2, \dots, M_p$) that achieves the following objectives:
\begin{enumerate}
	\item The difference between the lowest test accuracy of the models in the family and the test accuracy of $T$ as formulated below shall be minimized:  
	 $$|min\{TestAcc(M_1), \dots, TestAcc(M_p)\} - TestAcc(T)|$$
	
	\item For a given adversarial example $x$, the total number of models (in the family) successfully attacked by $x$ shall be minimized.
	\item The family size $p$ shall be maximized. 
\end{enumerate}

By trying to achieve all three objectives together, we may confront a situation similar to the space-time tradeoff. When we generate many models (increasing $p$), those models should be diverse enough such that the majority of them are not successfully attacked by the same adversarial example. 
Moreover, they cannot be too diverse as we need to achieve the legitimate-behavior preservation. 
Our idea exploration in solving this problem suggests that we may focus only on achieving the first two objectives and then we can keep increasing more models as long as we do not compromise (much) the first two objectives.

In the rest of this section, we first introduce a new metric, \textit{diversity measurement} (Section~\ref{subsec:diversity}), for the second objective, to measure the majority-model robustness. Then we explain the two components in our general framework along with empirical exploration on design choices for our model-generation technique (Section~\ref{sec:framework}).

\subsection{Diversity Measurement}
\label{subsec:diversity}
We measure the diversity of the model family based on the robustness of majority of models against adversarial examples generated targeting on one of the models. The detailed steps are as follows:
(1) generate adversarial examples for each model in the family; 
(2) for each generated example $i$, count the number of models that the example can successfully attack, denoted as $s_i$; 
(3) plot the distribution of $s_i$, i.e., the number of models that each example successfully attacks. The diversity of the model family can be calculated as
 $1 - \frac{\sum_i s_i}{a \times n}$, where $a$ is the number of adversarial examples and $n$ is the size of the model family.
Informally if most examples successfully attack fewer models in the family (low $s_i$), the model family is more diverse (close to $1$).

\subsection{General Defense Framework and Model-Generation Technique}
\label{sec:framework}
According to Figure~\ref{fig:toolframework}, given a target model, the model generator constructs a family of models. 
Models in the family (except the target model) are trained with additional adversarial examples from the specified attack approach. 
Note that the specified attack approach used to train the adversarial examples do not need to be the same as the attack approach that we are defending against. 

\subsubsection{Model Generator and Model-Generation Technique}

To generate other additional models in the family, we initially start with the simple idea of adversarial training~\cite{goodfellow2014explaining,Carlini2017Towards}.
In particular, given a target model $T$, we initially construct only one additional model $M_1$ that has the same architecture and parameter setting as $T$. 
However, $M_1$ is trained with an augmented dataset (the training set plus some adversarial examples for $T$).
Note that the adversarial examples are generated in the white-box attack scenario. Let $Adv_{M_i}$ denote the set of adversarial examples for $M_i$ and $Adv_T$ denote the set of adversarial examples (generated by the specified attack approach in the white-box attack scenario) for the target model $T$.
We notice that $M_1$ performs better on the adversarial examples for $T$ ($Adv_T$), but $M_1$ still performs worse on its own adversarial examples ($Adv_{M_1}$).
In fact, augmenting the training set cannot really improve the model against adversarial attacks.
Next, we construct more models, $M_1, M_2, M_3, \dots, M_p$, where $p$ denotes the number of additional models. The next question is how to train $M_2, M_3, \dots, M_p$.

For our model-generation technique, we devise the following two mechanisms to create the training set for each model:

\begin{itemize}
    \item \textbf{Solution 1.} The training set of $M_i$ is constructed as the union of the original training set and the adversarial examples generated for the previously constructed model $Adv_{M_{i-1}}$.
    \item \textbf{Solution 2.} The training set of $M_i$ is constructed as the union of the original training set and the adversarial examples generated for each of all the previously constructed models $Adv_{T}, Adv_{M_1}, Adv_{M_2}, \dots, Adv_{M_{i-1}}$.
\end{itemize}

\begin{figure}[t]
    \centering
    \includegraphics[scale=0.15]{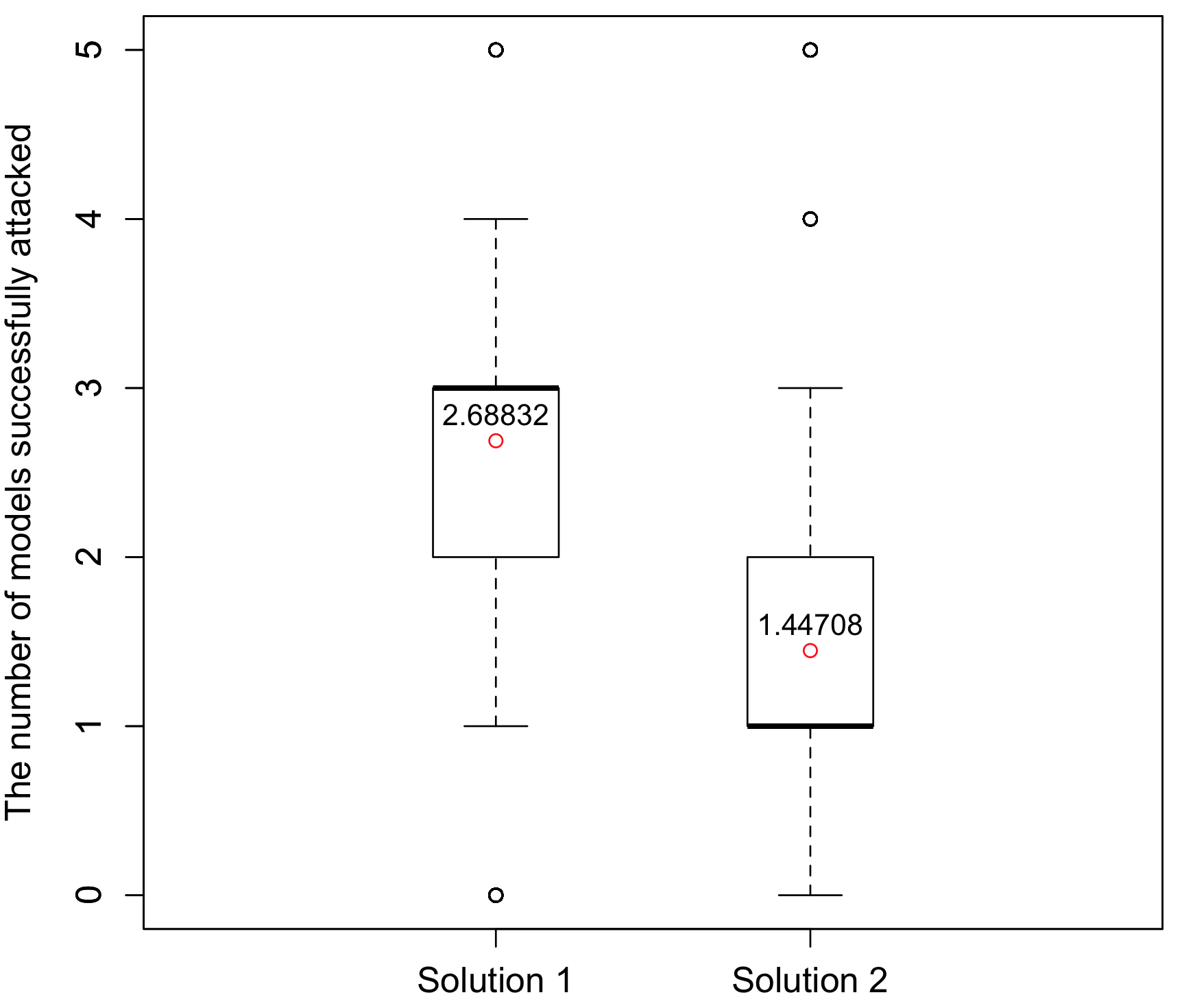}
    \caption{Boxplot comparison of the diversity measurement of two model families constructed by \textbf{Solution 1} and \textbf{Solution 2}. Each red dot represents the mean.}
    \label{fig:distribution-sol1-sol2}
\end{figure}

To compare the two solutions, we measure the diversity of two families of  models (with the family size chosen as 5), each of which is constructed by using the two proposed solutions. We find that a model family constructed by \textbf{Solution 2} performs better than that constructed by \textbf{Solution 1}. Figure~\ref{fig:distribution-sol1-sol2} shows the diversity measurement of the two model families. In \textbf{Solution 1}, the majority of the adversarial examples successfully attack 2 to 3 models. However, the majority of the adversarial examples successfully attack at most one model in \textbf{Solution 2}. Our experimental results also confirm that the model family constructed by \textbf{Solution 2} has a higher accuracy.

One observation to explain such results is that in the first solution, $Adv_{M_2}$ may not be representative for $Adv_{M_1}$. So $M_3$ (trained with the union of the original training set and $Adv_{M_{2}}$)  can still be vulnerable to $Adv_{M_{1}}$.
In the second solution, we train $M_3$  with the union of the original training set and $Adv_{T}, Adv_{M_1}, Adv_{M_2}$ to make $M_3$ more robust against all the previously constructed models' adversarial examples.
Thus, we implement \toolName by following the second solution.
It is worth noting that the last model, which is trained with all the other models' adversarial examples, seems to be more robust than other models.
However, the reason that we still include other models in our defense is  that the last model is vulnerable to its own adversarial set ($Adv_{M_3}$), and all the models in the family could be complementary to each other.

\begin{figure}[t]
    \centering
    \includegraphics[scale=0.25]{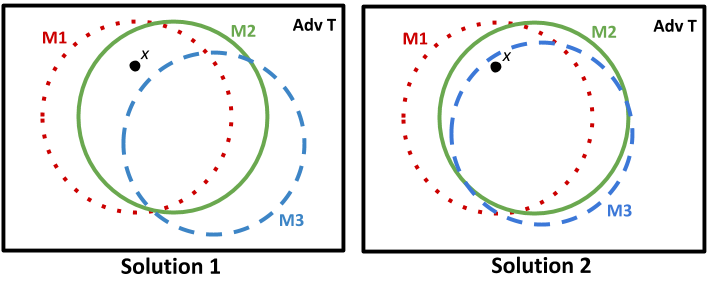}
    \caption{Comparison of the two solutions of augmenting the training set for constructing additional models.}
    \label{fig:approachintuition}
\end{figure}

Figure~\ref{fig:approachintuition} illustrates the idea of having multiple models and why \textbf{Solution 2} performs better than \textbf{Solution 1}.
Let the rectangle in each solution represent the set of all adversarial examples for the target model ($Adv_T \subset \mathbb{S}$) under a given attack approach. This set can be infinite. \toolName incrementally constructs additional models one by one.
First, \toolName constructs $M_1$ aiming to defend against a subset of $Adv_T$ by training $M_1$ with some adversarial examples for $T$. The circle for $M_1$ covers the subset of $Adv_T$ that $M_1$ can defend. \toolName keeps constructing more models to cover more space in the rectangle, indicating that \toolName is getting more robust to adversarial examples. Intuitively, \toolName performs well when many of the constructed models cover a large portion of the rectangle.
The difference between \textbf{Solution 1} and \textbf{Solution 2} is that \toolName trains $M_3$ with the union of the original training set and $Adv_T, Adv_{M_2}$ in \textbf{Solution 1}, but trains $M_3$ with the union of the original training set and $Adv_T, Adv_{M_1},Adv_{M_2}$ in \textbf{Solution 2}.
Thus, $M_3$ in \textbf{Solution 2} is likely able to defend against adversarial examples for $M_1$, resulting in having a higher chance that a given adversarial example can be defended by all the three models ($M1, M2, M3$).
According to Figure~\ref{fig:approachintuition}, a given adversarial example $x$ can be defended by only $M1$ and $M2$ in \textbf{Solution 1}. However, $x$ can be defended by all the three models in \textbf{Solution 2}.
A higher number of additional models that are robust to $x$ results in a higher chance that \toolName selects a right model at runtime.

\subsubsection{Runtime Model Selector}

To combine multiple models together, \toolName randomly selects a model from the family of models $T, M_1, M_2, M_3, \dots, M_p$ to compute the class label for each given input example.
The intuition of this strategy is to be able to introduce uncertainty in the target model (by the design principle of diversity) within the family of models so that it is hard for the attackers to generate adversarial examples that can attack all or most of the models.
Moreover, this runtime model selector acts as a wall to hide the gradient of a single model, because the attackers do not know in advance which model \toolName ends up selecting at runtime.
This random strategy shares the same spirit of defensive distillation~\cite{Nicolas2016Distillation}, which attempts to hide the gradient of the target model.

\begin{figure}[t]
\centering
  \begin{subfigure}[b]{0.35\textwidth}
    \includegraphics[width=\textwidth]{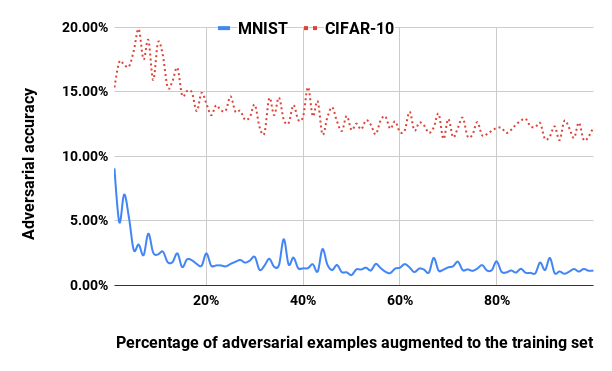}
    \caption{AdvAcc of target model $T$ when its training set is augmented with different quantities of adversarial examples.}
    \label{fig:adversarial-training1}
  \end{subfigure}
  \begin{subfigure}[b]{0.35\textwidth}
    \includegraphics[width=\textwidth]{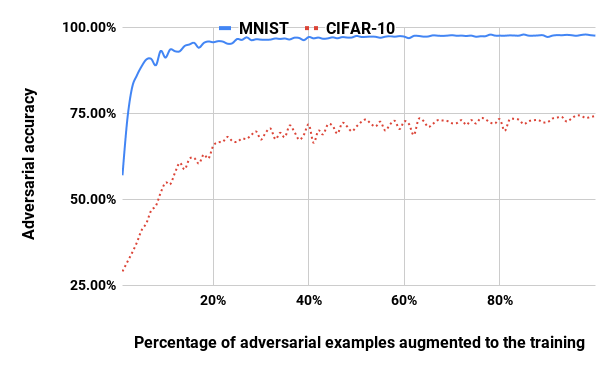}
    \caption{AdvAcc of model $D$ (on $Adv_{T}$) when its training set is augmented with different quantities of $Adv_{T}'$.}
    \label{fig:adversarial-training2}
  \end{subfigure}
  \caption{Results of using adversarial training against FGSM white-box attack for both datasets.}
\end{figure}

\subsubsection{Empirical Exploration on Design Choices}

Before settling down on our approach, we also conduct some  experiments to see whether using only adversarial training can make the target model more robust against adversarial examples.
We create two convolutional neural network models that can achieve about 99.13\%/80.4\% test accuracy (on the test set) for both MNIST and CIFAR-10.
Then we use FGSM to attack the model in the white-box attack scenario.
Without adversarial training, the model has about 8.87\%/13.79\% adversarial accuracy against FGSM for MNIST/CIFAR-10.
Then we try augmenting the training set with adversarial examples (generated by FGSM) to see whether the model is more robust.
Note that we run the experiment three times and report the average accuracy.
Figure~\ref{fig:adversarial-training1} shows that augmenting adversarial examples for both datasets can even worsen the model: the more adversarial examples we augment, the lower adversarial accuracy the model achieves.
According to Figure~\ref{fig:adversarial-training1}, the adversarial accuracy of target model $T$ for MNIST/CIFAR-10 surprisingly goes down to under 2.50\%/13.00\% when we augment the training set with more than 50\% of the original training set's size.

How about we construct another model $D$ that is robust to the target model's adversarial examples ($Adv_{T}$)?
Thus, we try to construct a new model $D$ with the same architecture as $T$, and train $D$ with the original training set augmented with a different set of adversarial examples for $T$ ($Adv_{T}'$) to see how $D$ performs on $Adv_{T}$.
The results in Figure~\ref{fig:adversarial-training2} show that the adversarial accuracy of model $D$ on $Adv_{T}$ is higher when we augment more adversarial examples especially in the beginning for both datasets.
Then the adversarial accuracy converges to around 97\%/74\% for MNIST/CIFAR-10.
Notice that the adversarial accuracy does not significantly change when we augment more than around 15\% and 25\% of adversarial examples for MNIST and CIFAR-10, respectively. 
So we decide that in our \toolName approach, we construct other models by using 15\% and 25\% of adversarial examples to augment the training set for MNIST and CIFAR-10, respectively.

One may wonder why we decide to augment the training set with adversarial examples.
We can also use only adversarial examples for $T$ to train other additional models.
If we use only adversarial examples to train other additional models, the additional models will perform worse on legitimate examples (in the test set), not being desirable.

According to the preceding observation, simply retraining the target model with adversarial examples cannot significantly make the model more robust against future adversarial examples as the attack approach also knows everything about the retrained target model.
However, having multiple models helps as an adversarial example for one model may not be able to fool another model.
The runtime model selector helps combine multiple models together. We use the random selection strategy because it is simple, low-cost, and provides some probabilistic guarantee that \toolName will not likely select a model that is vulnerable to a given adversarial example when we have many models.
Our evaluation (described in the next two sections) includes experiments to investigate on how many additional models our model-generation technique can generate while achieving legitimate-behavior preservation and robust diversity. 

%% file: evaluationsetup.tex
\section{Evaluation Setup}
\label{sec:experimentsetup}
We discuss the datasets used to evaluate our \toolName defense approach, against the three existing attack approaches.
Note that we run all the evaluations three times and report the average accuracy to reduce chance of accidental observations.

\noindent \textbf{Datasets.}
We conduct our evaluations on two public datasets: MNIST and CIFAR-10 (described in Section~\ref{sec:background}).

\noindent \textbf{Target model.}
We create two different convolutional neural network models for the two datasets (MNIST and CIFAR-10) as follows:

\begin{itemize}
\item \textbf{MNIST}: We follow the previous study on FGSM~\cite{papernot2017cleverhans} and use the same model in the study to evaluate FGSM with MNIST.
The model mainly consists of three convolutional layers with 64 neurons for each layer and ReLU as the activation function, and a densely-connected layer of 10 neurons for each digit.
\item \textbf{CIFAR-10}: We slightly adjust the model for CIFAR-10 in the study by C\&W~\cite{Carlini2017Towards}.
The model mainly consists of four convolutional layers with 64 neurons for the first two layers, 128 neurons for the next two layers, and ReLU as their activation function, two densely-connected layers of 256 neurons, and a densely-connected layer of 10 neurons for each class.
The only difference is that we add dropouts in both the convolutional layers and densely-connected layers.
We also add l2-norm regularization in the first two densely-connected layers.
\end{itemize}

We set the max epoch equal to 10 for the MNIST model and 50 for the CIFAR-10 model.
Two training processes both adopt the early stopping technique used to avoid overfitting.
The technique stops the training process when the validation loss fails to reduce by at least 0.001 for 5 epochs.
The target model achieves 98--99\%/76--80\% test accuracy for MNIST/CIFAR-10.
%
%

\noindent \textbf{Attack approaches.}
To check whether a model outputs a correct label for an adversarial example generated by an attack approach, we need to set the parameters of the attack approach to constrain the amount of perturbation on legitimate examples, so that we can use the original label as the ground truth.

In FGSM, the degree of perturbation is controlled by the parameter \textit{eps}. We set \textit{eps} to be 0.3/0.05 for MNIST/CIFAR-10 for both white-box and black-box scenarios.

In C\&W, the degree of perturbation is controlled by the parameter \textit{confidence}.
We set this value to 0.01 for both MNIST and CIFAR-10 in the white-box attack scenario. In order to enhance the black-box attack, we increase this value to 30 for both MNIST and CIFAR-10.
Another parameter named \textit{max\_iterations} is used to control the max number of iterations for generating adversarial examples.
We set \textit{max\_iterations} to 1000 for both MNIST and CIFAR-10 because this value is high enough to generate adversarial examples for  reducing the target model's accuracy to 0\%.
Other parameters in C\&W are set as their default values.

In PGD, we set the parameter \textit{eps} the same as in the FGSM experiments because the PGD approach is built on top of FGSM.
We reduce \textit{eps\_iter} to 0.01 from its default value 0.05, since we set \textit{eps} to 0.05 for CIFAR-10.
We also increase \textit{nb\_iter} to 30 from its default value 10 for CIFAR-10.

\noindent \textbf{\toolName setup.}
There are two main parameters to configure \toolName: (1) the percentage of adversarial examples augmented to the training set, and (2) the number of additional models to be constructed.
As discussed in Section~\ref{sec:approach}, we select 15\%/25\% as the percentage of augmented adversarial examples for MNIST/CIFAR-10.
To explore an optimal number of additional models, we set the number of additional models to be 1, 2, 3, and 4. Thus in total, \toolName has at most 5 models including the target model.

%% file: evaluationresult.tex
\section{Evaluation Results}
\label{sec:evalresults}
We measure the effectiveness of \toolName against FGSM, C\&W, and PGD in the white-box and black-box attack scenarios using two datasets: MNIST and CIFAR-10.

\subsection{Effectiveness of \toolName in White-box Attack Scenario}

%
%



To evaluate our approach in the white-box attack scenario, we perform two types of attacks to \toolName:

\begin{enumerate}
	\item \textbf{Non-adaptive Attack.}
	Because \toolName constructs a family of models: $T, M_1, M_2, \dots$, we can try the three attacks (FGSM, C\&W, and PGD) in a divide-and-conquer fashion by attacking each model separately to come up with the strongest adversarial examples for \toolName.
	\item \textbf{Adaptive Attack.}
	Defending against non-adaptive attacks is necessary but not sufficient. Thus, we introduce two adaptive attacks built on top of FGSM and PGD.
\end{enumerate}


\noindent \textbf{Effectiveness of \toolName under Non-adaptive Attack.}
For non-adaptive attack, we generate a set of adversarial examples by using the existing attacks for each model in the model family to find the strongest adversarial set (the one with the highest attack success rate). We measure the adversarial accuracies of \toolName along with each model in the model family against all sets of adversarial examples. The adversarial accuracy of \toolName that we report is the lowest one (which is based on the strongest possible adversary) among all models.
The median adversarial $L_{\infty}$ distance from FGSM and PGD attacks for MNIST/CIFAR-10 is $0.3/0.05$. The median adversarial $L_2$ distances for C\&W+MNIST, C\&W+CIFAR-10 are $1.829$ and $0.1823$, respectively.

\begin{figure}[t]
 \includegraphics[width=0.35\textwidth]{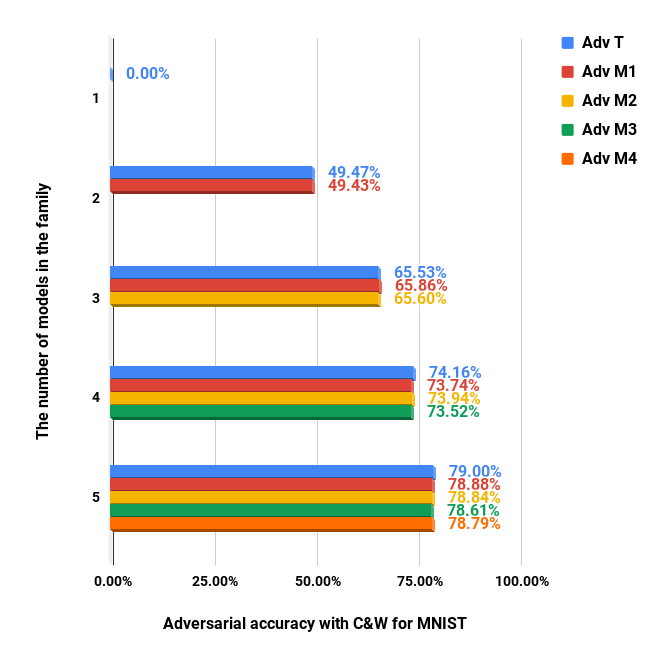}
 \vspace{-3ex}
 \caption{Adversarial accuracy of \toolName against C\&W on different sets of adversarial examples for MNIST.}
 \label{fig:overallresult-cw-mnist}
\end{figure}

\begin{figure}[t]
 \includegraphics[width=0.35\textwidth]{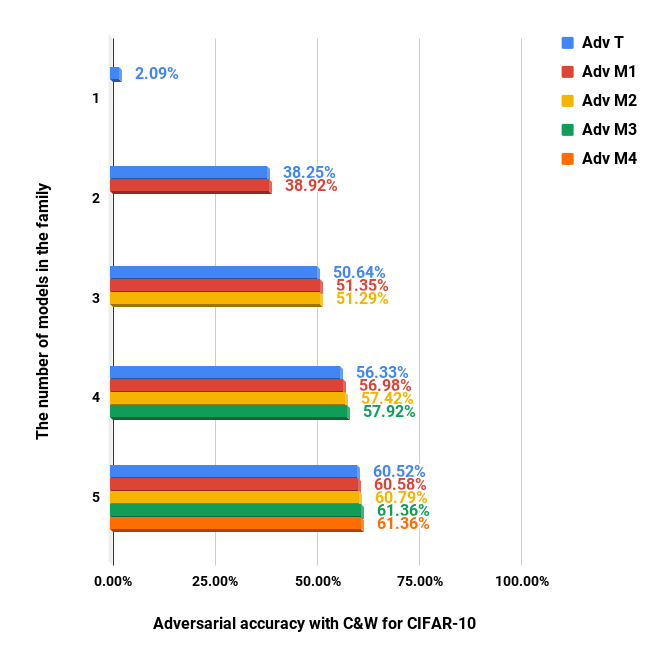}
 \vspace{-3ex}
 \caption{Adversarial accuracy of \toolName against C\&W on different sets of adversarial examples for CIFAR-10.}
 \label{fig:overallresult-cw-cifar}
\end{figure}

Figures~\ref{fig:overallresult-cw-mnist} and ~\ref{fig:overallresult-cw-cifar} show the adversarial accuracy of the target model $T$ (\toolName with only one model -- the target model) and the adversarial accuracy of \toolName (with 2, 3, 4, and 5 models) on different sets of adversarial examples ($Adv_T, Adv_{M_1}, \dots, Adv_{M_4}$ denoted as different colored bars).
Due to space limit, we show the  figures of results only for CW (the state-of-the-art attack).

The figures show that the more models \toolName has in the family, the higher adversarial accuracy \toolName gains.
Here we present the case only when \toolName has 5 models.
The results for the other attacks also follow the same trend. \textbf{In summary, the adversarial accuracies in FGSM+MNIST, FGSM+CIFAR-10, C\&W+MNIST, C\&W+CIFAR-10, PGD+MNIST, and PGD+CIFAR-10 are 64.12\%, 49.99\%, 78.61\%, 60.52\%, 73.51\%, 55.91\%, respectively, whereas the target model (without any defense) has 11.46\%, 14.47\%, 0.00\%, 2.09\%, 0.01\%, and 15.20\% adversarial accuracy for those setups, respectively.} These results indicate that \toolName can successfully defend \textbf{Non-adaptive Attack}.

In addition, we also compare our  \toolName approach with some existing defense approaches. The adversarially trained network proposed by Madry et al.~\cite{madry2018towards} has improved the adversarial accuracy over 90\% for MNIST against PGD,  outperforming our approach. However, for CIFAR-10, their approach and our approach have about the same accuracy of around 50--60\%, but we use a higher  \textit{eps} (0.05 vs. 0.03). 
Na et al.~\cite{na2018cascade} proposed another kind of adversarial training, \textit{cascade adversarial training}, which transfers the
knowledge of the end results of adversarial training. Their approach performs better for MNIST as their approach can reach the adversarial accuracy of 81--97\%. Our approach performs better for CIFAR-10 as their approach reaches only the accuracy of 27--38\%. We suspect that for complex images like CIFAR-10, our approach can outperform  the existing defense approaches.

\begin{table}[t] 
	\caption{Comparison of the adversarial accuracy of \toolName against \textbf{Non-adaptive attack}, \textbf{Adaptive Attack}, and \textbf{Gradient Obfuscation} for MNIST and CIFAR-10.}
	\label{tbl:adaptive}
	\vspace{-2ex}
\small
\renewcommand{\arraystretch}{1} \centering
\begin{tabular}{*{1}{|>{\centering\arraybackslash}p{0.15\columnwidth}}*{4}{|>{\centering\arraybackslash}p{0.1\columnwidth}}*{1}{|>{\centering\arraybackslash}p{0.19\columnwidth}}|}
\hline \multirow{2}{*}{\textbf{Datasets}}   & \multicolumn{2}{c|}{\textbf{Non-adaptive}}   &
\multicolumn{2}{c|}{\textbf{Adaptive}} & \textbf{Gradient} \\ \cline{2-5} & \textbf{FGSM} & \textbf{PGD} & \textbf{FGSM} & \textbf{PGD} & \textbf{Obfuscation}\\ \hline
MNIST           & 64.12\%      & 73.51\%         & 65.30\%      & 22.40\%	& 48.31\%    \\ \hline
CIFAR-10        & 49.99\%      & 55.91\%         & 35.27\%      & 22.69\%	& 35.30\%  \\ \hline
\end{tabular}
\end{table}

\noindent \textbf{Effectiveness of \toolName under Adaptive Attack.} 
To extensively evaluate our \toolName approach, we develop two adaptive attacks built upon FGSM and PGD. These two adaptive attacks are designed specifically to attack our approach. Because our approach randomly selects a model to predict for a given example, we can improve FGSM and PGD (which are made for attacking a single model) to consider all the models in the model family during their generation of adversarial examples. Instead of maximizing the loss function of a single model (Section ~\ref{sec:background}), we maximize the sum of all the loss functions as follows:
$$
x' = x + \epsilon \cdot \textit{sign}(\nabla_x \sum_{i} J_i(x, l))
$$
where $J_i$ is the loss function of each model $i$ in the model family. We conduct experiments on our  \toolName approach with 5 models for both MNIST and CIFAR-10 datasets. We also test our adaptive attack when the size of the perturbation is unbounded (setting the \textit{eps} to be $1.0$ for MNIST), and the adaptive attack built upon PGD can reach about 97\% attack success rate (i.e., \toolName has only 3\% accuracy). Thus, this adaptive attack is effective enough. Table~\ref{tbl:adaptive} shows the adversarial accuracy of \toolName against adaptive FGSM and adaptive PGD.
We can see that \toolName has relatively low adversarial accuracy against \textbf{Adaptive Attack}, but it still provides substantial defense for the target model. We also compare \toolName with a baseline attack approach presented in previous work about the gradient obfuscation attack~\cite{icml2018athalye}, which is used to attack stochastic gradients in randomized defenses. Compared with this baseline attack approach, \toolName achieves higher accuracy against \textbf{Gradient Obfuscation}, denoted as the last column in Table~\ref{tbl:adaptive}. For all the adaptive attacks (including Gradient Obfuscation), the median adversarial $L_{\infty}$ distance for MNIST/CIFAR-10 is $0.3/0.05$.

\subsection{Cross-Attack Scenario}
In the white-box attack scenario, we construct models in \toolName to defend against one attack approach by using adversarial examples generated based on the same attack approach.
We see that \toolName performs very well. Thus, is \toolName attack-dependent?
Realistically, we cannot know in advance which white-box attack approach the attackers will use in reality, indicating that we cannot construct models in \toolName based on the adversarial examples generated by the approach to be used by the attackers.
Therefore, we further investigate the robustness of our approach in the cross-attack scenario, in which we launch one attack approach (e.g., C\&W) to test \toolName built on top of adversarial examples of a different attack approach (e.g., FGSM).

Table~\ref{tbl:crossattack} shows that no matter which group of adversarial examples \toolName uses to construct models, \toolName performs better than the target model.
Surprisingly, we find that when being attacked by C\&W, \toolName built on FGSM adversarial examples is more robust than it built on C\&W adversarial examples.
Note that it does not necessarily suggest that FGSM attack is always more powerful than C\&W but instead suggests that FGSM adversarial examples are more suitable than C\&W ones to construct diversified models in \toolName.

\begin{figure}[t]
    \centering
    \includegraphics[scale=0.18]{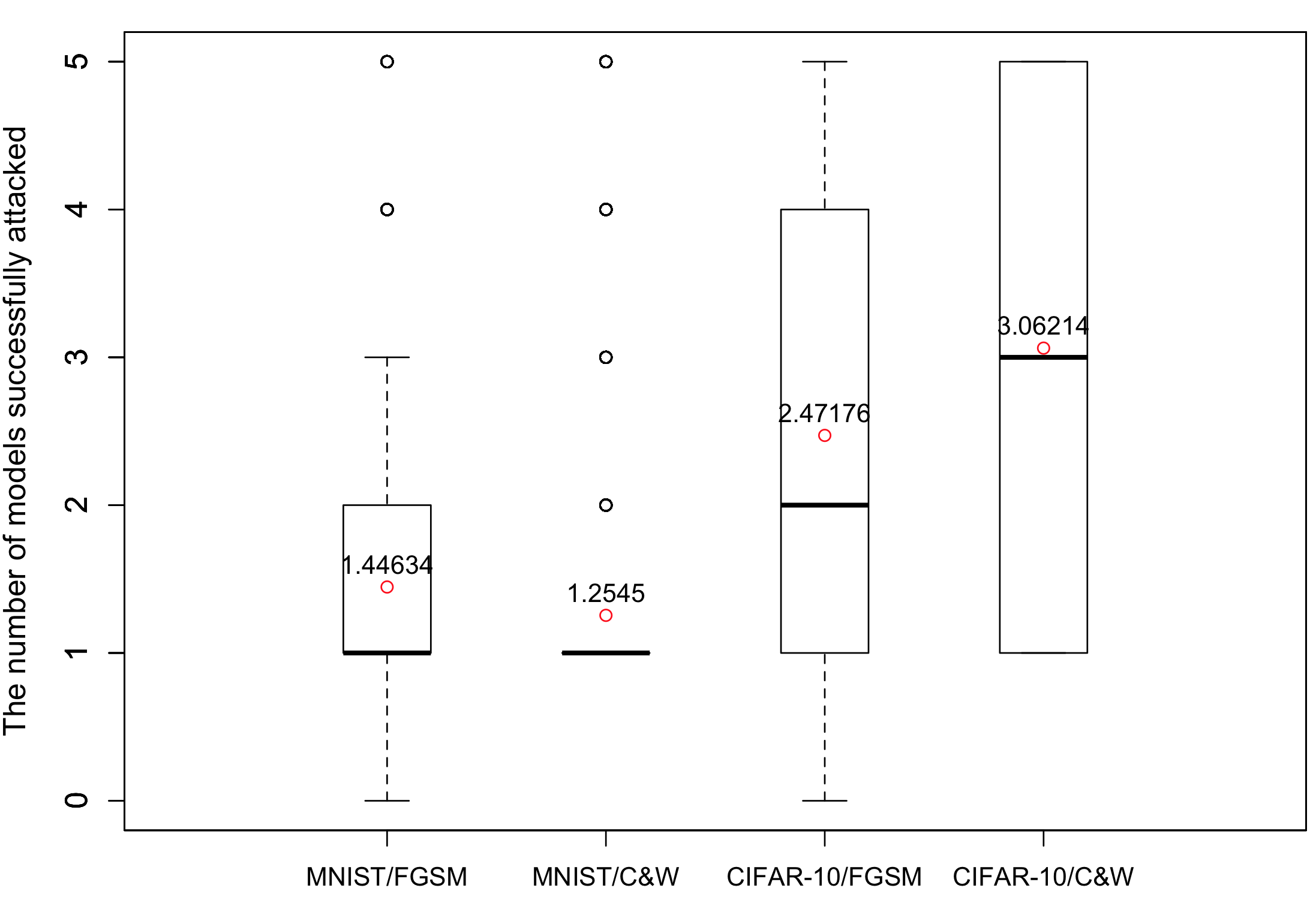}
    \caption{Comparison of the diversity measurement of four model families against FGSM and C\&W for MNIST and CIFAR-10. Each red dot represents the mean.}
    \label{fig:distribution-cross-attack}
\end{figure}

\begin{table}[t]
	\caption{Adversarial accuracies of \toolName built on FGSM and C\&W adversarial examples against FGSM and C\&W white-box attacks. Cross-attack results are marked in \textbf{bold}.}
	\vspace{-2ex}
	\label{tbl:crossattack}
\small
	\renewcommand{\arraystretch}{1.2}
    \centering
    \begin{tabular}{*{1}{|>{\centering\arraybackslash}p{0.1\columnwidth}}*{1}{|>{\centering\arraybackslash}p{0.18\columnwidth}}*{1}{|>{\centering\arraybackslash}p{0.155\columnwidth}}*{2}{|>{\centering\arraybackslash}p{0.155\columnwidth}}|}
        \hline
        \multicolumn{2}{|c|}{\multirow{2}{*}{\textbf{Attacks}}} & \multirow{2}{*}{Tgt model}  & \multicolumn{2}{c|}{\toolName} \\ \cline{4-5}
        \multicolumn{2}{|c|}{} & & FGSM adv exps & C\&W adv exps \\ \hline
        \multirow{2}{*}{\textbf{FGSM}}  & MNIST     & 11.46\%   & 64.12\%  &  \textbf{14.96\%}   \\ \cline{2-5}
        \multirow{2}{*}{}               & CIFAR-10  & 14.47\%   & 49.99\%  &  \textbf{26.47\%}   \\ \hline
        \multirow{2}{*}{\textbf{C\&W}}  & MNIST     & 00.00\%   & \textbf{69.19\%}  & 49.44\%    \\ \cline{2-5}
        \multirow{2}{*}{}               & CIFAR-10  & 00.00\%   & \textbf{61.25\%}  & 60.45\%    \\ \hline

    \end{tabular}
\end{table}
We further measure the diversity of model families in \toolName (defined in Section~\ref{subsec:diversity}) built on FGSM adversarial examples for both datasets.
In diversity measurement, we use adversarial examples generated with both FGSM and C\&W attacks. Figure~\ref{fig:distribution-cross-attack} shows 
the distribution of adversarial examples by the number of models that each example successfully attacks in four model families (each model family containing five models).
We observe that although previous results suggest that \toolName built on FGSM examples is more effective, the diversity of the model family for CIFAR-10 (CIFAR-10/FGSM) is not very high ($1-\frac{2.47}{5} = 0.51$). The average number of models that can be attacked in the model family is 2.47.
One possible reason is that the data in CIFAR-10 have high dimensions (RGB images with a lager number of pixels). Small changes in each dimension would enable the adversarial examples to be adapted to the diverse models.
Model families trained for MNIST present more robustness diversity than model families trained for CIFAR-10.

\subsection{Effectiveness of \toolName on Original Test Dataset}

While achieving higher adversarial accuracy, \toolName also maintains about the same test accuracy as the target model.
Table~\ref{tbl:overallresult-test} shows that in each model family, the additional models have about the same test accuracy as the target model.
Overall, against FGSM, \toolName for MNIST/CIFAR-10 has 98.91\%/77.13\% test accuracy compared to 99.03\%/79.78\% (the target model's test accuracy in column T).
Against C\&W, \toolName for MNIST/CIFAR-10 has 98.79\%/80.05\% test accuracy compared to 98.58\%/80.41\%. Against PGD, \toolName for MNIST/CIFAR-10 has 99.03\%/73.18\% test accuracy compared to 99.24\%/76.20\%. These results show that \toolName can maintain similar accuracy on testing examples (the original test dataset).

\begin{table}[t]
	  \caption{Test accuracy of each model in \toolName with 5 models: $T$ (target model)$, M_1, M_2, M_3, M_4$.}
	\label{tbl:overallresult-test}
	\vspace{-2ex}
\footnotesize
  \renewcommand{\arraystretch}{1.2}
    \centering
    \begin{tabular}{*{1}{|>{\arraybackslash}p{0.25\columnwidth}}*{5}{|>{\centering\arraybackslash}p{0.07\columnwidth}}|}
        \hline

        \multirow{2}{*}{\textbf{Attack / dataset}}   & \multicolumn{5}{c|}{\textbf{Test acc of each model (\%)}}  \\ \cline{2-6}

        & \textbf{$T$} & \textbf{$M_1$} & \textbf{$M_2$} & \textbf{$M_3$} & \textbf{$M_4$}   \\ \hline
        \textbf{FGSM / MNIST}      & 99.03 & 98.65 & 98.90 & 98.91 & 98.93 \\ \hline
        \textbf{FGSM / CIFAR-10}   & 79.78 & 76.58 & 76.89 & 76.46 & 76.69 \\ \hline
        \textbf{C\&W / MNIST}      & 98.98 & 99.10 & 99.04 & 99.04 & 99.05 \\ \hline
        \textbf{C\&W / CIFAR-10}   & 77.52 & 76.64 & 76.29 & 75.34 & 75.83 \\ \hline
        \textbf{PGD / MNIST}       & 99.24 & 99.10 & 98.99 & 99.01 & 98.97 \\ \hline
        \textbf{PGD / CIFAR-10}    & 76.20 & 73.90 & 72.41 & 72.72 & 71.50 \\ \hline
    \end{tabular}
\end{table}

\subsection{Impact of Different Numbers of Models}

Figures~\ref{fig:overallresult-cw-mnist} and ~\ref{fig:overallresult-cw-cifar} show that having more models in \toolName can increase adversarial accuracies.
Surprisingly, each model can maintain about the same test accuracy as the target model's.
One may have the concern that when we generate more models, the last model may have lower test accuracy than the others, because the last model is constructed by the most adversarial examples augmented to its training set.
It turns out that those adversarial examples do not have a negative impact or confuse the model when the model faces against legitimate examples.
This result suggests that having more models could make the classifier more robust against adversarial examples.

\begin{figure}[t]
  \includegraphics[width=0.45\textwidth]{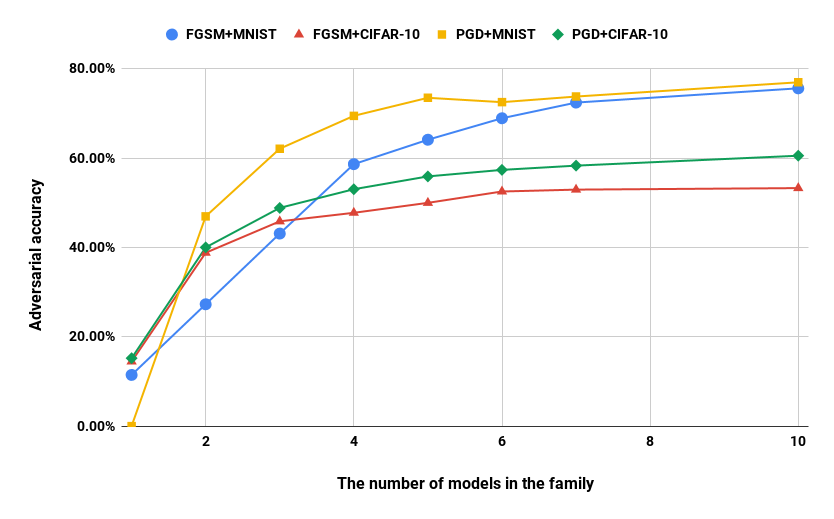}
  \vspace{-4ex}
  \caption{Adversarial accuracy of \toolName with different numbers of models based on \textbf{Non-adaptive Attack}.}
  \label{fig:differentnums}
\end{figure}

Statistically, having more models increases the chance of selecting a robust model for each given input example at runtime. This assumption is true only when additional models are complementary to the existing models. In other words, the diversity of the model family is increasing. Thus we additionally evaluate our approach when using more models. Figure~\ref{fig:differentnums} shows the white-box adversarial accuracy of \toolName.
The adversarial accuracies converge when using 7--10 models. The result implies that at some point, adding more models does not increase robustness diversity. Ideally, it is better to have more models to improve robustness diversity. So these results open more future research directions on how to improve the model generator to be more effective in generating complementary models.

\subsection{Impact of Adversarial Training and Randomization in \toolName}

\begin{table*}[t]
	 \caption{Three adversarial accuracies on a set of adversarial examples generated
		for each model in \toolName with 5 models ($T, M_1, M_2, M_3, M_4$) against
		an attack approach and a dataset. For the set of adversarial examples for model $x$ ($Adv_{x}$), (1)
		the \textbf{first} accuracy denotes the accuracy of model $x$ on $Adv_{x}$, (2)
		the \textbf{second}  accuracy denotes the average accuracy of all the models constructed (by \toolName) \textbf{before} model $x$ on $Adv_{x}$, and (3) 
		the \textbf{third} accuracy denotes the average accuracy of all the models constructed (by \toolName) \textbf{after} model $x$ on $Adv_{x}$. Note that \toolName constructs $M_1, M_2, M_3, $ and $M_4$ in order. The highest accuracy among the three is in bold.}
	\label{tbl:impact-adv-random}
	\vspace{-2ex}
\footnotesize
	\renewcommand{\arraystretch}{1.2}
    \centering
    \begin{tabular}{*{1}{|>{\centering\arraybackslash}p{0.24\columnwidth}}*{1}{|>{\centering\arraybackslash}p{0.29\columnwidth}}*{3}{|>{\centering\arraybackslash}p{0.29\columnwidth}}*{1}{|>{\centering\arraybackslash}p{0.29\columnwidth}}|}
        \hline

        \multirow{2}{*}{\textbf{Atk / dataset}}   & \multicolumn{5}{c|}{\textbf{Adversarial accuracies on adversarial examples}}  \\ \cline{2-6}

        & \textbf{$Adv_T$} & \textbf{$Adv_{M_1}$} & \textbf{$Adv_{M_2}$} & \textbf{$Adv_{M_3}$} & \textbf{$Adv_{M_4}$}   \\ \hline
        \textbf{FGSM / MNIST}      & 11.46\% / - / \textbf{95.40\%}  & 01.94\% / 54.42\% / \textbf{95.93\%} & 04.10\% / 66.78\% / \textbf{95.74\%} & 04.10\% / 76.34\% / \textbf{96.48\%}  & 04.73\% / \textbf{80.54\%} / -    \\ \hline
        \textbf{FGSM / CIFAR-10}   & 14.47\% / - / \textbf{61.08\%}  & 14.62\% / 57.10\% / \textbf{65.29\%} & 13.71\% / 63.01\% / \textbf{68.43\%} & 19.27\% / 59.75\% / \textbf{68.49\%}  & 20.85\% / \textbf{57.46\%} / -   \\ \hline
        \textbf{C\&W / MNIST}      & 00.00\% / - / \textbf{98.57\%}  & 00.00\% / 98.18\% / \textbf{98.58\%} & 00.00\% / 98.32\% / \textbf{98.67\%} & 00.00\% / 98.03\% / \textbf{98.75\%}  & 00.00\% / \textbf{98.07\%} / -    \\ \hline
        \textbf{C\&W / CIFAR-10}   & 02.09\% / - / \textbf{74.58\%}  & 03.33\% / \textbf{75.47\%} / 74.50\% & 03.53\% / \textbf{75.72\%} / 74.57\% & 05.54\% / \textbf{75.32\%} / 74.73\%  & 07.55\% / \textbf{74.94\%} / -    \\ \hline
        \textbf{PGD / MNIST}       & 00.01\% / - / \textbf{97.03\%}  & 00.00\% / 94.72\% / \textbf{97.46\%} & 00.04\% / 92.85\% / \textbf{97.16\%} & 00.55\% / 91.74\% / \textbf{96.54\%}  & 00.71\% / \textbf{91.64\%} / -    \\ \hline
        \textbf{PGD / CIFAR-10}    & 15.20\% / - / \textbf{65.81\%}  & 15.06\% / 65.37\% / \textbf{66.18\%} & 17.92\% / 65.93\% / \textbf{66.49\%} & 18.71\% / 66.96\% / \textbf{67.01\%}  & 16.63\% / \textbf{66.63\%} / -    \\ \hline
    \end{tabular}
   
\end{table*}


Based on the idea of adversarial training, the model generator in \toolName constructs each additional model one after another, where $M_i$ is trained with the union of the original training set and $Adv_{T}, Adv_{M_1}, Adv_{M_2}, \dots, Adv_{M_{i-1}}$.
In other words, for $Adv_{M_i}$, all the other models constructed after $M_i$ are trained with the training set that includes $Adv_{M_i}$.
Therefore, most models constructed after $M_i$ should be robust to $Adv_{M_i}$.
To test this hypothesis, we measure the average adversarial accuracy (on $Adv_{M_i}$) of all the models constructed after $M_i$, denoted as the third accuracy in Table~\ref{tbl:impact-adv-random}. We can see that the third accuracy is often the highest as expected. There are a few cases that the third accuracy is not the highest but is similar to the second accuracy for \textbf{C\&W / CIFAR-10}. We inspect these cases and find that the third, fourth, and fifth models are not more resilient than the second model in the family, indicating that only two models in the family already reach the saturation point of adversarial accuracy. Thus, we would need to improve the model construction in our approach in order to reach higher adversarial accuracy for \textbf{C\&W / CIFAR-10}.
Note that the cells in the last column in Table~\ref{tbl:impact-adv-random} are missing the third accuracy, because $M_4$ is the last model in the family.

In order to select a robust model by the runtime model selector with a high chance, most models constructed before $M_i$ should be robust to $Adv_{M_i}$ as well. Thus, we measure the average accuracy of all the models constructed before $M_i$ on $Adv_{M_i}$, denoted as the second accuracy in Table~\ref{tbl:impact-adv-random}. Our result shows that the second accuracy is also relatively high compared to the accuracy of $M_i$ on $Adv_{M_i}$ (denoted as the first accuracy in the table). This result raises a question on why
most models constructed before $M_i$ can achieve higher accuracy on $Adv_{M_i}$, even though
they are not exposed to $Adv_{M_i}$ at all during their training process.
One possible reason is that $Adv_{M_i}$ is very specific to $M_i$, especially for C\&W adversarial examples. And every new model constructed is trained with more adversarial examples, causing the later constructed models to be different from previously constructed models.




\subsection{Effectiveness of \toolName in Black-box Attack Scenario}

For the black-box attack scenario where the attackers can access only the target model output, we first train a substitute model with synthetic inputs selected by a Jacobian-based heuristic~\cite{Nicolas2017Practical} to approximate the target model's decision boundaries.
We use 150 hold-out images from the test set and run 5 Jacobian-based augmentation epochs, and set the augmentation parameter $\lambda$ = 0.1. All of these parameters are default values. Then we apply white-box attacks on the substitute model to generate adversarial examples and evaluate the target model on those examples.

Table~\ref{tbl:whiteblackresult} shows that \toolName still achieves higher adversarial accuracy than the target model in the black-box attack scenario except when C\&W and CIFAR-10 are used. Reasons that our defense does not substantially improve the target model in the black-box attack scenario might be the transferability property of adversarial examples or the fact that black-box attack approaches are not powerful enough to reveal weakness in the target model.

%

\begin{table}[t]
	 \caption{Comparison of adversarial accuracy of the target model and \toolName in the black-box attack scenario.}
	\vspace{-2ex}
	\label{tbl:whiteblackresult}
\footnotesize
	\renewcommand{\arraystretch}{1.2}
    \centering
    \begin{tabular}{*{1}{|>{\centering\arraybackslash}p{0.12\columnwidth}}*{1}{|>{\centering\arraybackslash}p{0.2\columnwidth}}*{1}{|>{\centering\arraybackslash}p{0.18\columnwidth}|>{\centering\arraybackslash}p{0.18\columnwidth}}|}
        \hline
        \multicolumn{2}{|c|}{\multirow{2}{*}{\textbf{Attacks}}} & \multicolumn{2}{c|}{\textbf{Black-box}} \\ \cline{3-4}
        \multicolumn{2}{|c|}{} & Tgt model & \toolName \\ \hline
        \multirow{2}{*}{\textbf{FGSM}}  & MNIST     & 70.19\%   & \textbf{81.76\%}     \\ \cline{2-4}
        \multirow{2}{*}{}               & CIFAR-10  & 70.32\%   & \textbf{72.82\%}     \\ \hline
        \multirow{2}{*}{\textbf{C\&W}}  & MNIST     & 72.10\%   & \textbf{78.89\%}     \\ \cline{2-4}
        \multirow{2}{*}{}               & CIFAR-10  & \textbf{70.71\%} & 65.28\%       \\ \hline

    \end{tabular}
\end{table}

%% file: discussion.tex
\section{Discussion}

In our experiments, we also try a number of different target models for MNIST. Most of them achieve about the same accuracy around 98--99\%; however, we notice that some of the target models are more robust than the others as they can weaken the transferability property of the adversarial examples. We find that if we use a more robust target model, our defense system can achieve substantially higher adversarial accuracy.


We choose the random strategy for runtime model selector to achieve low cost and  introduce uncertainty in the target model. Nevertheless, there can be another strategy based on multiple-model/implementation majority voting~\cite{pei2017deepxplore,srisakaokul2018:multitest}, which we plan to explore in future work. In particular, Srisakaokul et al.~\cite{srisakaokul2018:multitest} use multiple-implementation testing to test an implementation of a machine learning algorithm, where the majority output across multiple implementations of the same algorithm is used as a test oracle. \toolName also contains multiple models that are robust to a given adversarial example, so we may be able to use the majority label across multiple models as an output, instead of randomly selecting one model to be applied on the given example at runtime. However, as discussed in Section~\ref{sec:intro}, such strategy can be costly, needing to apply all models in the family to the given example. In addition, with a candidate adversarial example, the attackers can also simulate such strategy of multi-model majority voting ahead of time and know beforehand which model in the family to (re-)compute the gradient to improve the candidate adversarial example if needed.

%% file: relatedwork.tex
\section{Related Work}

A variety of approaches have been proposed for defending against adversarial examples.
Meng et al.~\cite{Meng2017MagNet} proposed a defense approach named MagNet against adversarial examples. MagNet consists of two main steps: detect and reform. Like \toolName, MagNet does not modify the target model. MagNet first detects whether a given input example is adversarial by measuring the distance between the input example and the manifold of legitimate examples in the training set. If the input example is farther to the manifold of the legitimate examples, the input example is marked as an adversarial example and then gets reformed/reconstructed to be close to legitimate examples. Their ideas are different from our \toolName approach, because \toolName does not reform the given example. Instead, \toolName trains more models to handle any example. MagNet can perform well in a \textit{gray-box} attack scenario (where the attackers know about the target model and the defense, but not the parameters of the defense), but cannot handle a white-box attack scenario at all. In fact, MagNet achieves only less than 20\%/40\% adversarial accuracy against FGSM/C\&W attack (in the white-box attack scenario) for MNIST~\cite{defensegan}. So our approach outperforms MagNet.


Xu et al.~\cite{Xu:NDSS18} proposed an approach to detect an adversarial example by ``squeezing'' out unnecessary input features. Then the approach compares the model's prediction on the original example with its prediction on the squeezed example. If the predictions are substantially different, the original example is likely to be adversarial. However, their approach does not produce any output if the given input is adversarial. 

Rouhani et al.~\cite{Rouhani2018DeepFense} proposed an approach of using complementary but disjoint modular redundancies to defend against adversarial examples. Although Rouhani et al. claim their approach to be attack-independent, it still requires a lot of pre-constructed modules, which are needed to be trained with different attack algorithms. Their approach only guarantees to detect whether a given input is a legitimate input or not, but there is no guarantee on the accuracy of the output when the input is adversarial.

Song et al.~\cite{song2018pixeldefend} proposed an approach of image purification named PixelDefend to defend against adversarial examples. PixelDefend requires no knowledge of the attack approach nor the target model, but uses the PixelCNN model for its state-of-the-art performance in modeling image distributions~\cite{Oord:2016} to detect an adversarial example. Then PixelDefend purifies the adversarial example by searching for more probable images within a small distance of the adversarial example. Because PixelDefend only purifies the given adversarial example, we can combine this approach with \toolName to proceed with the purified example.

Tramer et al.~\cite{tramer2018ensemble} proposed another version of adversarial training, named Ensemble Adversarial Training, which augments the training set with adversarial examples crafted from pre-trained models. Our model generator also constructs each additional model with the original training set augmented with adversarial examples generated from other previously constructed models in the family. However, all the models in the family have the same architecture and parameter settings, whereas the pre-trained models in Ensemble Adversarial Training can have different architectures or parameter settings.


Zantedeschi et al.~\cite{Zantedeschi:2017} proposed an approach to make the target model more robust against adversarial examples by reinforcing the model architecture so that its prediction becomes more stable. Their approach uses the Bounded ReLU activation function for hedging against the forward propagation of adversarial perturbation and Gaussian data augmentation during training. Their approach is mainly for making the attack visually detectable. However, it still does not perform well against C\&W.

%% file: conclusion.tex
\section{Conclusion}
In this paper, we have proposed \toolName, a defense approach against adversarial examples for neural networks. Our approach consists of (1) a general defense framework based on multiple models and (2) a technique for generating these multiple models to achieve high defense capability. In particular, we construct a family of models such that the models are complementary to each other to accomplish robustness diversity. 
Our approach is simple, scalable, and easy to be applied, because it does not modify the target model. We evaluate our approach on three attack strategies (FGSM, C\&W, and PGD) for two datasets (MNIST and CIFAR-10). The evaluation results show that our defense approach substantially improves the target model's adversarial accuracy by 22--74\% in the white-box attack scenario against both attack strategies and both datasets. Even for the black-box attack scenario, our defense approach also improves the target model's adversarial accuracy by 2--10\%. While making the target model more robust against adversarial examples, \toolName still maintains similar accuracy as the target model on legitimate examples.